%% file: main.tex
\documentclass{article}

\PassOptionsToPackage{numbers,sort&compress}{natbib}
\usepackage[preprint]{neurips_2026}
\usepackage[T1]{fontenc}
\usepackage{microtype}
\usepackage{amsmath,amssymb}
\usepackage{graphicx}
\usepackage{booktabs}
\usepackage{capt-of}
\usepackage{wrapfig}
\usepackage{xcolor}
\usepackage[colorlinks,allcolors=blue!45!black]{hyperref}

\newcommand{\method}{LoopCache}
\newcommand{\leftcaption}[1]{\caption[#1]{\raggedright #1}}
\let\cite\citep

\title{\method{}: Accelerating Long-Context Self-Speculative Decoding for Looped Transformers}
\author{%
  Zizhuo Fu \qquad Haochen Huang \qquad Shengxuan Qiu \qquad Linye Wei\\
  \textbf{Ziyu Tang \qquad Runsheng Wang \qquad Meng Li}%
  \thanks{Contact email: \href{mailto:fuzizhuo@stu.pku.edu.cn}{fuzizhuo@stu.pku.edu.cn}.
    Corresponding author: Meng Li (\href{mailto:meng.li@pku.edu.cn}{meng.li@pku.edu.cn}).}\\
    \\
  Peking University
}

\begin{document}

\maketitle

\input{docs/1-abstract}
\input{docs/2-introduction}
\input{docs/3-background}
\input{docs/4-motivation}
\input{docs/5-method}
\input{docs/6-experiment}
\input{docs/7-conclusion}

\bibliographystyle{abbrvnat}
\bibliography{main}

\end{document}

%% file: docs/1-abstract.tex
\begin{abstract}
Looped Transformers achieve large effective depth with compact models by repeatedly applying a weight-shared block across $R$ recurrent steps, each maintaining a loop-specific key-value (KV) cache.
To accelerate sequential loop execution, recent wave-style self-speculative methods like WaveFront Decoding (WFD) and LoopSpec pipeline draft and target tokens across loops in a single step.
However, they face two critical limitations.
First, they draft only at fixed loops and ignore intermediate prediction changes, wasting computation on stale trajectories.
Second, batching tokens from different loops loads up to $R\times$ the KV cache per step, severely amplifying memory traffic in long-context scenarios.
To address these challenges, we propose \method{}, an efficient self-speculative decoding framework for long-context Looped Transformers.
Based on the correlation between prediction switches and output confidence, \method{} introduces confidence-guided dynamic draft updating to refine drafts in flight across all intermediate loops.
Furthermore, \method{} employs cyclic loop KV cache scheduling that rotates through loop-specific KV caches, restricting per-step KV traffic to a single loop while batching tokens at the nearest matching cycle for lossless verification.
By also selectively retaining independent KV caches for a few loop-sensitive shallow layers, \method{} improves acceptance rates with negligible overhead.
Experiments show that \method{} achieves average 10.25$\times$ and 14.56$\times$ speedup over WFD and LoopSpec.
\mbox{Code repository: \href{https://github.com/zZzZ9zZ9/LoopCache}{https://github.com/zZzZ9zZ9/LoopCache}.}
\end{abstract}

%% file: docs/2-introduction.tex
\section{Introduction}
\label{sec:introduction}

\begin{wrapfigure}{r}{0.46\linewidth}
    \centering
    \includegraphics[width=\linewidth]{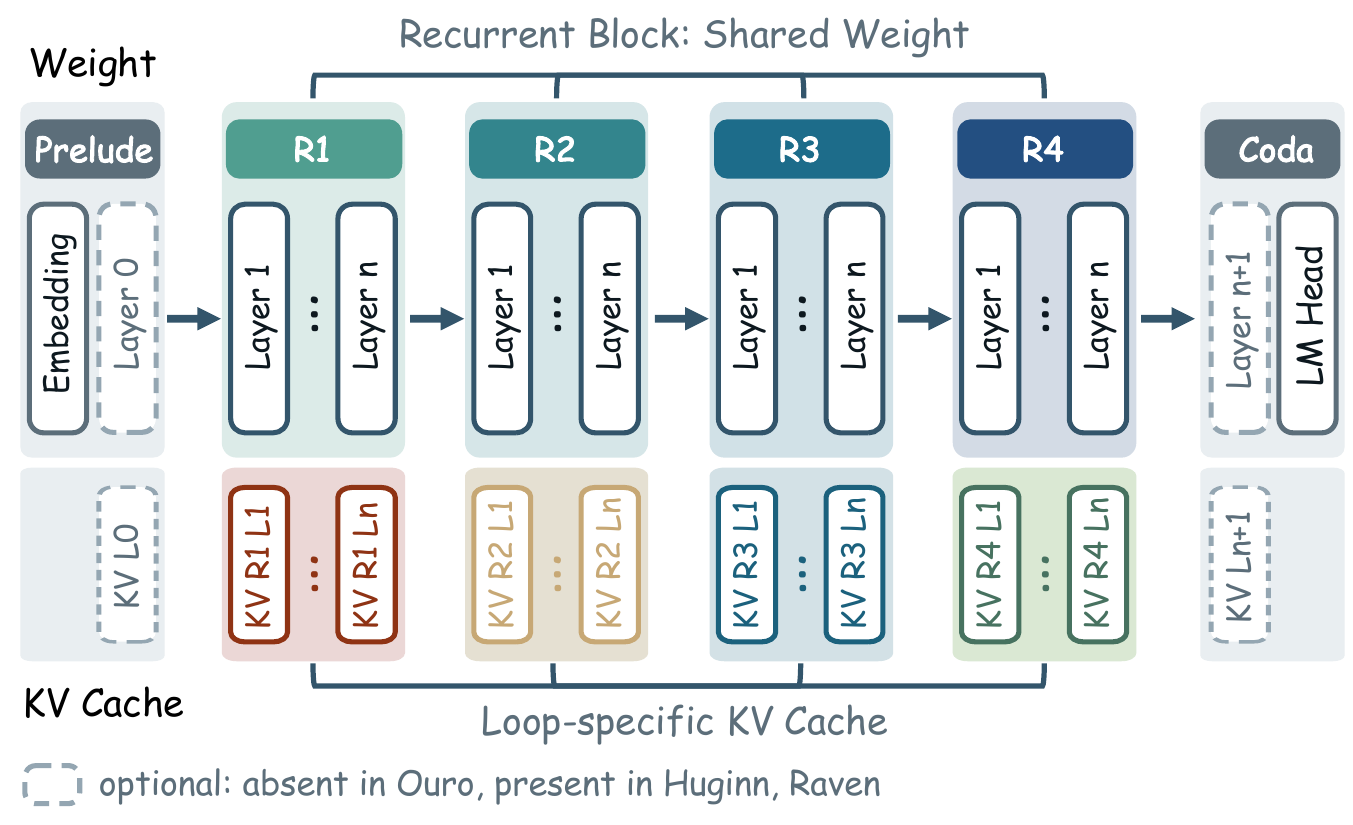}
    \caption{Architecture of looped Transformers.}
    \label{fig:loop_intro}
\end{wrapfigure}

Looped Transformers have recently emerged as a promising architecture for scaling test-time compute~\cite{Geiping2025ScalingUT, Zhu2025ScalingLR, McLeish2025TeachingPL,Yang2026Nanbeige423BUA,Prairie2026ParcaeSL,Kim2026ResidualQuantKC,Fu2025ThinkatHardDL}. By repeatedly applying a shared block of layers across multiple recurrent steps, these models refine hidden representations iteratively, increasing effective depth without adding parameters. 
Meanwhile, modern applications such as chain-of-thought reasoning~\cite{Chen2025TowardsRE, Ling2025LongReasonAS, Qiu2025PHYBenchHE} and long-horizon agents~\cite{Huang2026ASO, Yang2026Nanbeige423BUA} increasingly demand long-context decoding over tens of thousands of tokens~\cite{Bai2023LongBenchAB, Fu2025H2EALHA, Fu2026ScalingPA, Liu2024ClusterKVML, Chen2024ArkValeEG, Xu2025SpeContextEE,Fu2026AttentionSF}. However, combining recurrent depth with long contexts creates a severe inference bottleneck, as every generated token requires multiple sequential passes through the recurrent block while attending to \mbox{a massive key-value (KV) cache.}

\begin{wrapfigure}{r}{0.58\linewidth}
    \centering
    \includegraphics[width=\linewidth]{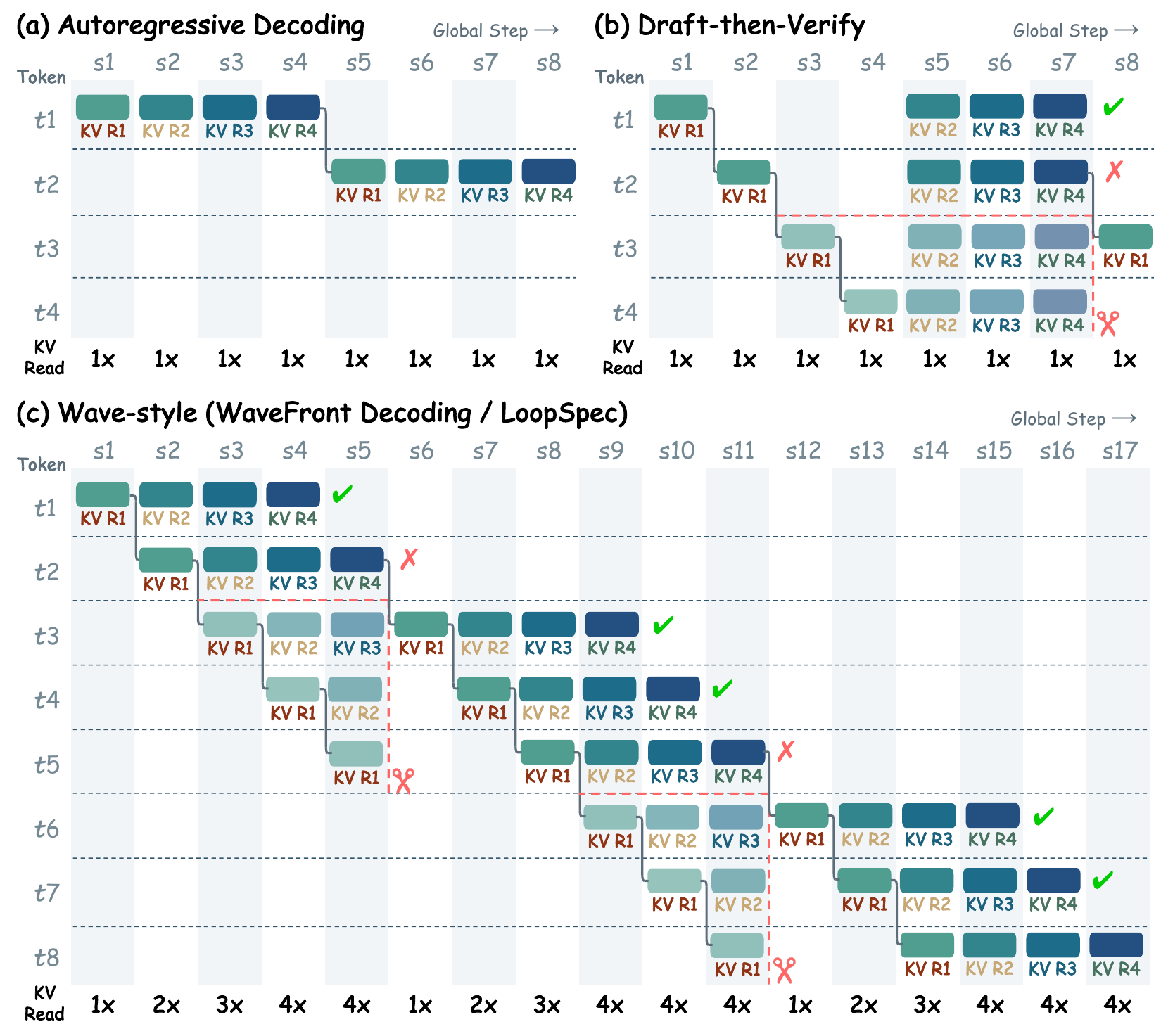}
    \caption{Token generation dataflows and per-step KV cache read overheads for a Looped Transformer ($R=4$) under (a) AutoRegressive, (b) Draft-then-Verify, and (c) WaveFront Decoding / LoopSpec.}
    \label{fig:loop_background}
\end{wrapfigure}

As illustrated in Fig.~\ref{fig:loop_intro}, a looped Transformer contains a recurrent block between optional prelude and coda layers. The block runs for $R$ loops, where $R$ denotes the recurrence depth. 
Although the recurrent steps share the same model weights, each step maintains its own KV cache. 
Under conventional autoregressive (AR) decoding, each token must complete all $R$ recurrent steps before the next token can be processed, as shown in Fig.~\ref{fig:loop_background}(a). 
Nevertheless, intermediate recurrent states can already provide useful next-token predictions, so a shallow loop can act as a natural draft model for the full-depth model. This makes looped transformers well suited to self-speculative decoding~\cite{Leviathan2022FastIF,Huang2026S2MoEEE,Elhoushi2024LayerSkipEE,zhang2024draft,Bang2026SpecMoEAF,Zhao2025AcceleratingLR,Sadhukhan2024MagicDecBT,Tiwari2025QuantSpecSD}.
A straightforward approach is Draft-then-Verify (DtV), shown in Fig.~\ref{fig:loop_background}(b). DtV first generates multiple draft tokens using shallow recurrent computation, then verifies them in a batch using the remaining recurrent steps.
However, drafting and verification remain separate phases, limiting their execution overlap.

To overcome the sequential bottleneck of DtV, recent works such as WaveFront Decoding (WFD)~\cite{Ha2026WaveFrontDP} and LoopSpec~\cite{Cho2026LoopSpecPS} introduce wave-style pipelined self-speculative decoding. Exploiting the shared weights across recurrent blocks, these methods batch draft and target tokens at different recurrent depths into a single global step, as shown in Fig.~\ref{fig:loop_background}(c). This wave-style pipeline enables early verification and immediate trajectory correction compared to DtV, achieving higher token throughput in short-context scenarios.

Despite this progress, existing wave-style methods face two critical challenges.
First, intermediate loops are underused. WFD and LoopSpec draft only at fixed loops, yet in a looped transformer every loop can produce a draft token.
The predicted next token may change between an early loop and an intermediate loop.
This change is ignored until the final loop verifies the token, and all later tokens built on the outdated draft are wasted.
Second, wave-style batching inflates KV cache traffic. The batched tokens in one step sit at different loops, and each token reads the KV cache of its own loop.
As Fig.~\ref{fig:loop_background}(c) shows, each step reads up to $R\times$ the KV cache of AR and DtV. In long-context scenarios, this $R\times$ memory traffic amplification completely offsets the benefits of pipelined early verification.

To address these challenges, we propose \method{}, an efficient self-speculative decoding framework for long-context looped transformers.
First, we observe that token predictions frequently change across recurrent loops, while predictions at later loops increasingly agree with the final token.
Based on this insight, \method{} introduces confidence-guided dynamic draft updating, which monitors intermediate loop predictions and dynamically updates draft tokens. 
Second, to eliminate redundant KV-cache accesses, we propose a cyclic loop KV cache scheduling strategy. 
Across global steps, \method{} accesses the loop-specific KV caches in a fixed cycle ($\text{KV R}_1$ to $\text{KV R}_R$), keeping the per-step KV cache access consistent with AR and DtV. To ensure lossless verification, tokens drafted under mismatched KV caches are batched at the nearest upcoming cycle with the matching loop KV cache to compute their exact target loops in parallel.
Furthermore, we identify a distinct layer-wise sensitivity across recurrent loops: shallow layers exhibit much larger KV cache variations across loops than deeper layers. By selectively retaining independent KV caches only for a small number of shallow layers during specific steps, \method{} further improves draft acceptance with~negligible~memory~overhead.

\begin{samepage}
\setlength{\leftmargini}{2em}
Our main contributions can be summarized as follows:
\begin{itemize}
    \item We analyze token switch behavior across recurrent loops and propose a confidence-guided dynamic draft updating mechanism. By flexibly leveraging predictions from all intermediate recurrent steps, \method{} dynamically refines draft tokens in flight to avoid \mbox{wasted computation on stale tokens.}
    \item Targeting long-context efficiency, we propose a cyclic loop KV-cache scheduling strategy which keeps per-step KV traffic consistent with AR and DtV, while batching exact verification at the nearest matching KV cycle. We further exploit layer-wise KV differences across loops by selectively decoupling shallow-layer KV caches to improve acceptance rates.
    \item Extensive experiments across multiple Looped Transformers and long-context benchmarks demonstrate that \method{} achieves average 10.25$\times$ and 14.56$\times$ speedup over WFD and LoopSpec, while delivering average 5.12$\times$ and 2.52$\times$ speedup compared to \mbox{standard AR and DtV decoding.}
\end{itemize}
\end{samepage}

%% file: docs/3-background.tex
\section{Background}
\label{sec:background}

\paragraph{Looped Transformers}
Looped Transformers extend recurrent computation to standard Transformer blocks by repeatedly applying a weight-shared block for $R$ steps~\cite{Dehghani2018UniversalT,Geiping2025ScalingUT, Zhu2025ScalingLR,McLeish2025TeachingPL,Yang2026Nanbeige423BUA,Prairie2026ParcaeSL}. As illustrated in Fig.~\ref{fig:loop_intro}, current architectures mainly follow two designs: Ouro~\cite{Zhu2025ScalingLR} repeats the entire Transformer stack, whereas Huginn~\cite{Geiping2025ScalingUT} and Raven~\cite{McLeish2025TeachingPL} iterate an internal recurrent block between a non-recurrent prelude and coda. Despite sharing model parameters across all recurrent steps, each loop $r \in \{1, \dots, R\}$ maintains an independent KV cache $\text{KV R}_r$ because its input hidden states evolve across loops~\cite{Geiping2025ScalingUT,Zhu2025ScalingLR}.

\paragraph{Self-Speculative Decoding}
Speculative decoding accelerates AR generation by using a fast draft model to propose candidate tokens and a target model to verify them in parallel, where rejection sampling guarantees an identical output distribution to the target model~\cite{Leviathan2022FastIF}. Self-speculative decoding eliminates the separate draft model by reusing intermediate computations of the target model itself~\cite{Elhoushi2024LayerSkipEE,zhang2024draft,Bang2026SpecMoEAF,Zhao2025AcceleratingLR,Sadhukhan2024MagicDecBT,Tiwari2025QuantSpecSD}. In Looped Transformers, early loops naturally serve as draft models for the full-depth model~\cite{Geiping2025ScalingUT,Zhu2025ScalingLR}. As shown in Fig.~\ref{fig:loop_background}(b) and Fig.~\ref{fig:loop_background}(c), while DtV serializes shallow drafting and deep verification, WFD~\cite{Ha2026WaveFrontDP} and LoopSpec~\cite{Cho2026LoopSpecPS} overlap these phases via pipelined execution, batching tokens at different recurrent depths into a single step. Specifically, WFD drafts at a predefined depth and organizes in-flight tokens into a diagonal wavefront. LoopSpec further introduces a primary proposal at depth $d_1$ and a gated residual proposal at a deeper depth $d_2$. However, this second proposal creates fallback branches that severely inflate the runtime batch size, reaching peak effective batch sizes of 176 and 191 on its two Raven models~\cite{Cho2026LoopSpecPS}. Moreover, both methods remain restricted to fixed drafting depths and access multiple loop-specific KV caches within each step.\looseness=-1

\paragraph{Cross-Loop KV Sharing}
To reduce KV cache overhead, several works share or reuse KV states across recurrent steps. Huginn explores a bounded cyclic cache that overwrites cache slots across loops~\cite{Geiping2025ScalingUT}, while Ouro~\cite{Zhu2025ScalingLR} and WFD~\cite{Ha2026WaveFrontDP} reuse the last-loop or averaged KV cache during decoding. Similarly, Mixture-of-Recursions (MoR) reuses the first-loop KV states across subsequent recursions~\cite{Bae2025MixtureofRecursionsLD}. Although these lossy approaches reduce memory footprint and KV traffic, they inevitably cause accuracy degradation. Specifically, KV states in later loops are iteratively refined, so reusing unrefined first-loop KV states severely compromises model quality. Conversely, reusing the last-loop KV cache is incompatible with standard training and prefill phases, where all tokens are processed in parallel and future-loop KV states are not yet available when computing early loops. This discrepancy prevents effective training adaptation and similarly degrades inference accuracy. In contrast to lossy KV sharing, \method{} reduces long-context KV traffic while preserving exact loop-specific verification for lossless acceleration.

%% file: docs/4-motivation.tex
\section{Motivation}
\label{sec:motivation}

\begin{figure}[!t]
    \centering
    \includegraphics[width=0.70\linewidth]{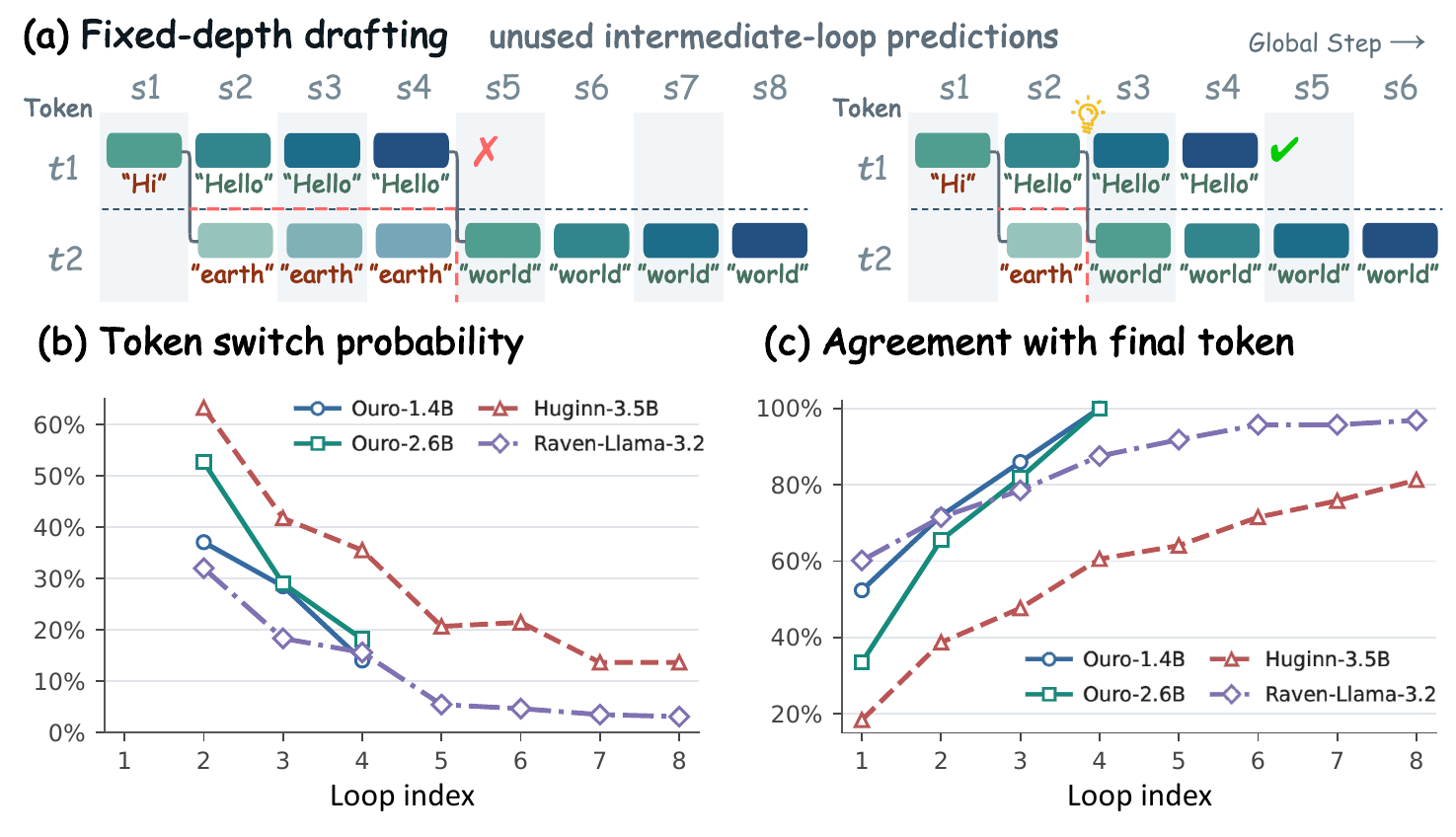}
    \caption{Intermediate-loop prediction behavior. (a) Fixed-depth drafting versus draft updating with intermediate-loop predictions. (b) Token switch probability across recurrent loops. (c) Agreement between intermediate-loop and final-loop predictions.}
    \label{fig:observation-1}

    \par\vspace{9pt}
    \begin{minipage}[t]{0.533\linewidth}
        \vspace{0pt}
        {\centering\includegraphics[height=0.30\textheight]{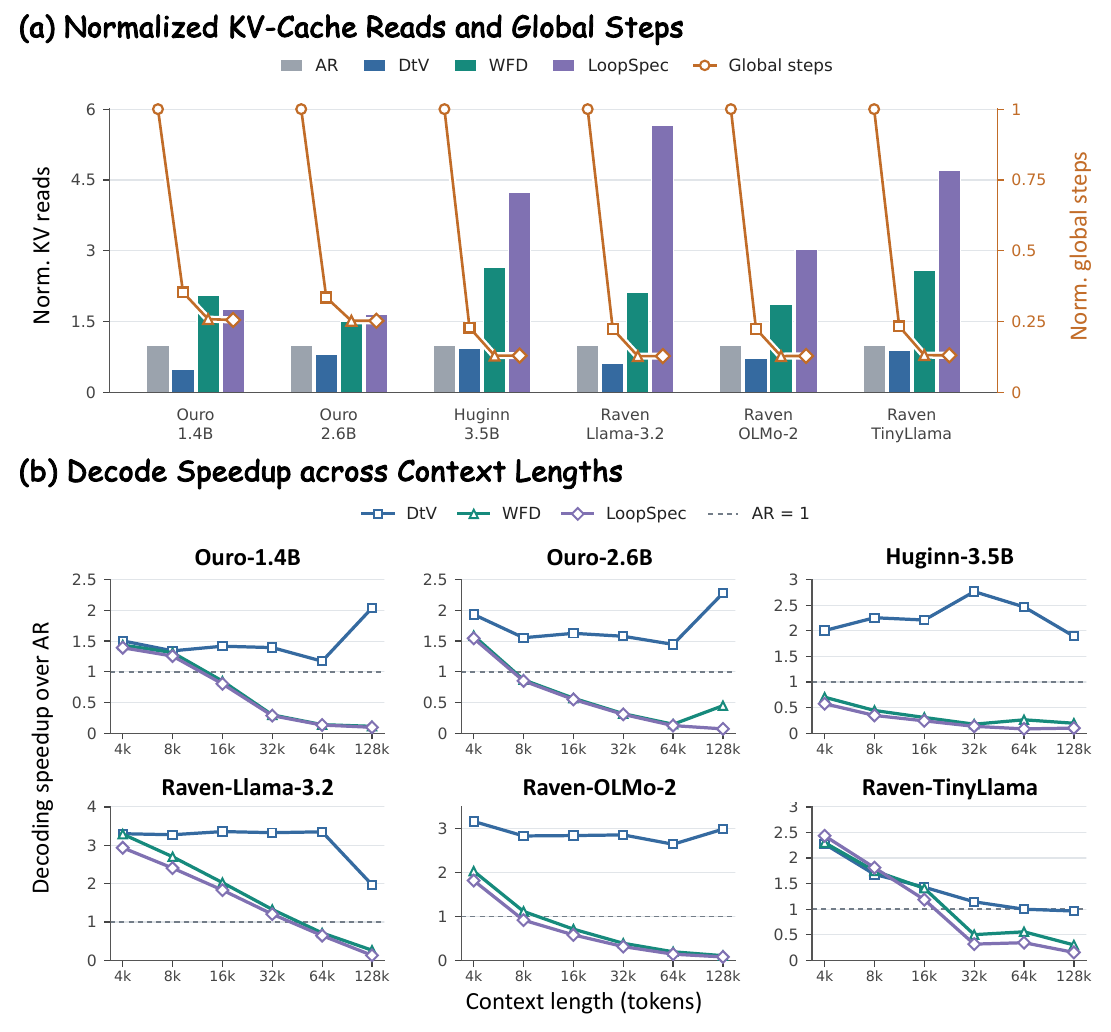}\par}
        \leftcaption{KV cache overhead and long-context decoding performance. (a) Total KV cache reads and global steps, normalized to AR. (b) Decoding speedup over AR across different context lengths.}
        \label{fig:observation-2}
    \end{minipage}\hfill
    \begin{minipage}[t]{0.442\linewidth}
        \vspace{0pt}
        {\centering\includegraphics[height=0.30\textheight]{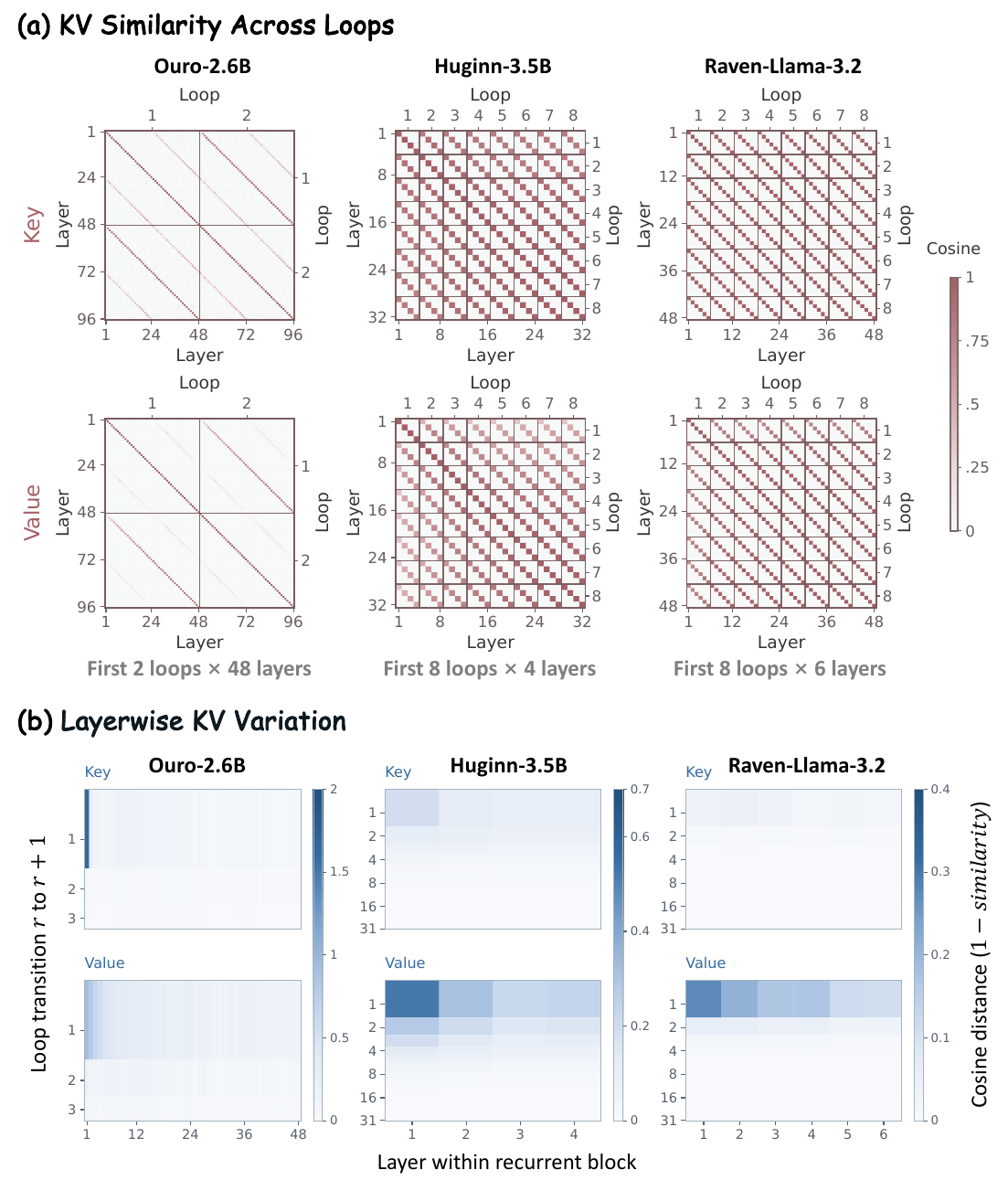}\par}
        \leftcaption{Cross-loop KV cache characteristics. (a) Pairwise cosine similarity of key and value states across recurrent loops. (b) Layerwise cosine distance of key and value states between consecutive loops.}
        \label{fig:observation-3}
    \end{minipage}
\end{figure}

\textbf{Observation 1: Intermediate-loop predictions enable dynamic draft refinement.}
Existing wave-style methods such as WFD and LoopSpec draft at fixed recurrent depths, leaving intermediate-loop predictions largely unused~\cite{Ha2026WaveFrontDP,Cho2026LoopSpecPS}.
However, every recurrent loop can produce a next-token prediction, which may change as the loop depth increases.
As illustrated in Fig.~\ref{fig:observation-1}(a), a draft token may change at an intermediate loop, while subsequent speculative tokens continue along the outdated trajectory.
Fig.~\ref{fig:observation-1}(b) shows that such prediction switches occur frequently, with distinct patterns across models and recurrent depths.
This variation highlights the limitations of fixed-depth drafting.
Although predictions from every loop could provide additional drafts, retaining all candidate branches can increase the batch size to $2^{R-1}$ in the worst case.
Such exponential growth is impractical for models with many recurrent loops.

Importantly, Fig.~\ref{fig:observation-1}(c) shows that agreement with the final prediction generally increases with loop depth, indicating that later predictions are more likely to match the final token.
This motivates dynamically replacing outdated drafts rather than retaining all candidate branches.
As shown on the right side of Fig.~\ref{fig:observation-1}(a), intermediate predictions can be used to update in-flight drafts and prune their outdated speculative continuations.
By guiding these updates with prediction confidence, \method{} exploits intermediate-loop information while maintaining a single active draft trajectory to~avoid~excessive~branching.

\textbf{Observation 2: KV cache traffic limits wave-style decoding in long contexts.}
WFD and LoopSpec reduce sequential execution by batching tokens at different recurrent depths, but each global step accesses multiple loop-specific KV caches~\cite{Ha2026WaveFrontDP,Cho2026LoopSpecPS}.
As shown in Fig.~\ref{fig:observation-2}(a), despite requiring fewer global steps than AR and DtV, both methods incur substantially higher total KV cache reads.
This overhead becomes increasingly critical as context length grows, since decoding becomes dominated by KV cache memory traffic.
Consequently, Fig.~\ref{fig:observation-2}(b) shows that WFD and LoopSpec suffer declining speedups as context length increases, often becoming slower than conventional AR decoding.

Directly sharing KV caches between draft and target loops cannot resolve this issue without compromising exact computation.
When a draft token is accepted, its recurrent computation must remain consistent with the target model.
However, using a mismatched loop KV cache during drafting produces incorrect hidden states and KV entries.
These states cannot be reused for subsequent target computation without recomputation, even if the draft token itself is accepted.
Therefore, lossless acceleration requires reducing KV traffic while preserving exact loop-specific computation during verification.
This motivates a cyclic KV cache access strategy that batches exact verification at the~corresponding~loop~cycles.

\textbf{Observation 3: Cross-loop KV similarity exhibits strong layer-wise differences.}
Although exact verification requires loop-specific KV states, the shared weights and iterative refinement of Looped Transformers suggest potential similarities across recurrent loops.
Fig.~\ref{fig:observation-3}(a) shows that the key and value states of corresponding layers exhibit high cosine similarity across loops.
This suggests that reusing KV caches from other loops during drafting may preserve useful intermediate predictions while reducing memory traffic.
Combined with exact loop-specific verification, such approximate reuse offers an opportunity to improve speculative decoding efficiency \mbox{without compromising correctness.}

However, cross-loop KV similarity varies considerably across layers.
As shown in Fig.~\ref{fig:observation-3}(b), shallow layers exhibit larger KV differences between consecutive loops, while deeper layers remain relatively stable.
Thus, uniformly sharing KV caches across all layers would degrade draft quality and reduce acceptance rates.
Instead, selectively maintaining independent KV caches for a few sensitive shallow layers can better preserve intermediate predictions with small memory overhead.
This provides an opportunity to improve draft acceptance while retaining the efficiency of cross-loop KV reuse.

%% file: docs/5-method.tex
\section{LoopCache}
\label{sec:method}

Building on the observations in Section~\ref{sec:motivation}, we propose \method{}, a self-speculative decoding framework that combines dynamic draft refinement with efficient KV cache scheduling.
As illustrated in Fig.~\ref{fig:method}, \method{} uses confidence-guided updates to maintain a single active draft trajectory and cyclically schedules loop-specific KV caches to reduce long-context memory traffic.
\mbox{A separate verification path} preserves exact target computation, while selective shallow-layer KV accesses further improve draft quality.

\subsection{Confidence-Guided Dynamic Draft Updating}

As discussed in Observation 1, existing wave-style methods draft at fixed recurrent depths, potentially wasting computation on outdated speculative trajectories~\cite{Ha2026WaveFrontDP,Cho2026LoopSpecPS}.
As shown in Fig.~\ref{fig:observation-1}(a), intermediate-loop predictions can correct these drafts before final verification.
However, unconditionally updating every changed prediction may cause repeated reversals, disrupting speculative progress.

A \textit{harmful switch} occurs when a replaced draft reappears at a later loop.
Let $z_r(v)$ denote the logit of token $v$ after loop $r$, and $y_r=\arg\max_v z_r(v)$ its prediction.
For a current draft $x$ and new candidate $s=y_r\neq x$, the harmful-switch indicator is
\begin{equation}
h_r(x\rightarrow s)=
\mathbf{1}\left[
\exists\, k\in\{r+1,\ldots,R\}:y_k=x
\right],
\label{eq:harmful_switch}
\end{equation}
where $\mathbf{1}[\cdot]$ is the indicator function.\\
The frequency of such reversals characterizes the harmful-switch risk.

\begin{wrapfigure}{r}{0.53\linewidth}
    \centering
    \includegraphics[width=\linewidth]{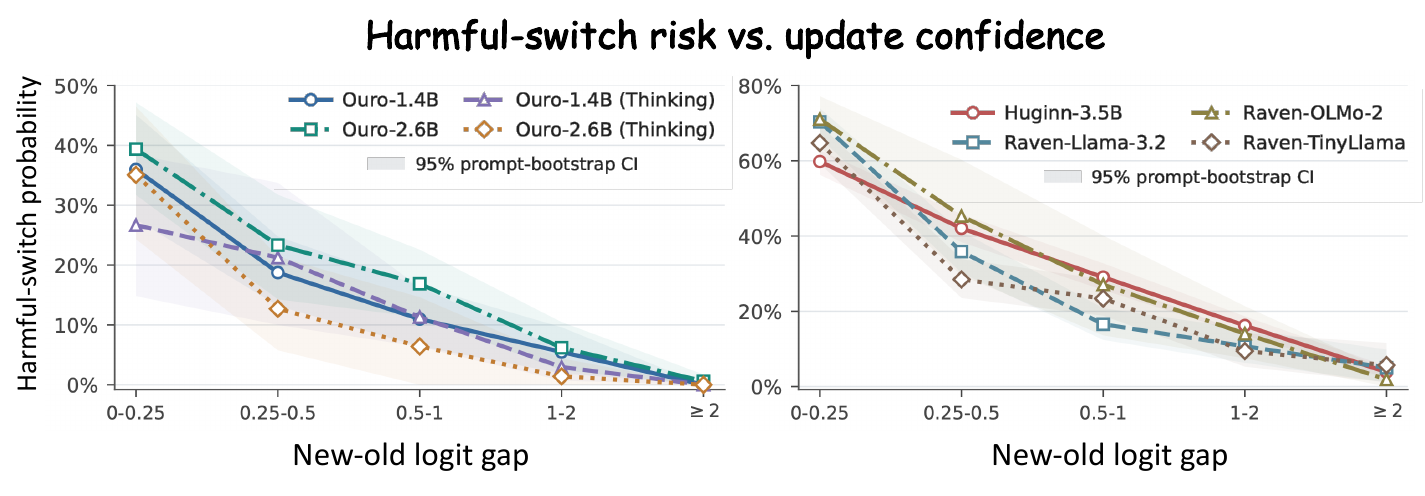}
    \caption{Relationship between harmful-switch probability and the new-old logit gap $\Delta z$ across different Looped Transformers.}
    \label{fig:method-observation-1}
\end{wrapfigure}

To avoid unnecessary switches, \method{} measures \textit{update confidence} using the new-old logit gap:
\begin{equation}
\Delta z_r
=z_r(s)-z_r(x)
=\log\frac{p_r(s)}{p_r(x)},
\label{eq:update_confidence}
\end{equation}
where $p_r$ denotes the softmax probability distribution.
This gap measures how strongly the current loop favors the new candidate over the existing draft.
Fig.~\ref{fig:method-observation-1} shows that harmful-switch probability decreases consistently as the logit gap increases across models.

\begin{figure}[!t]
    \centering
    \includegraphics[width=\linewidth]{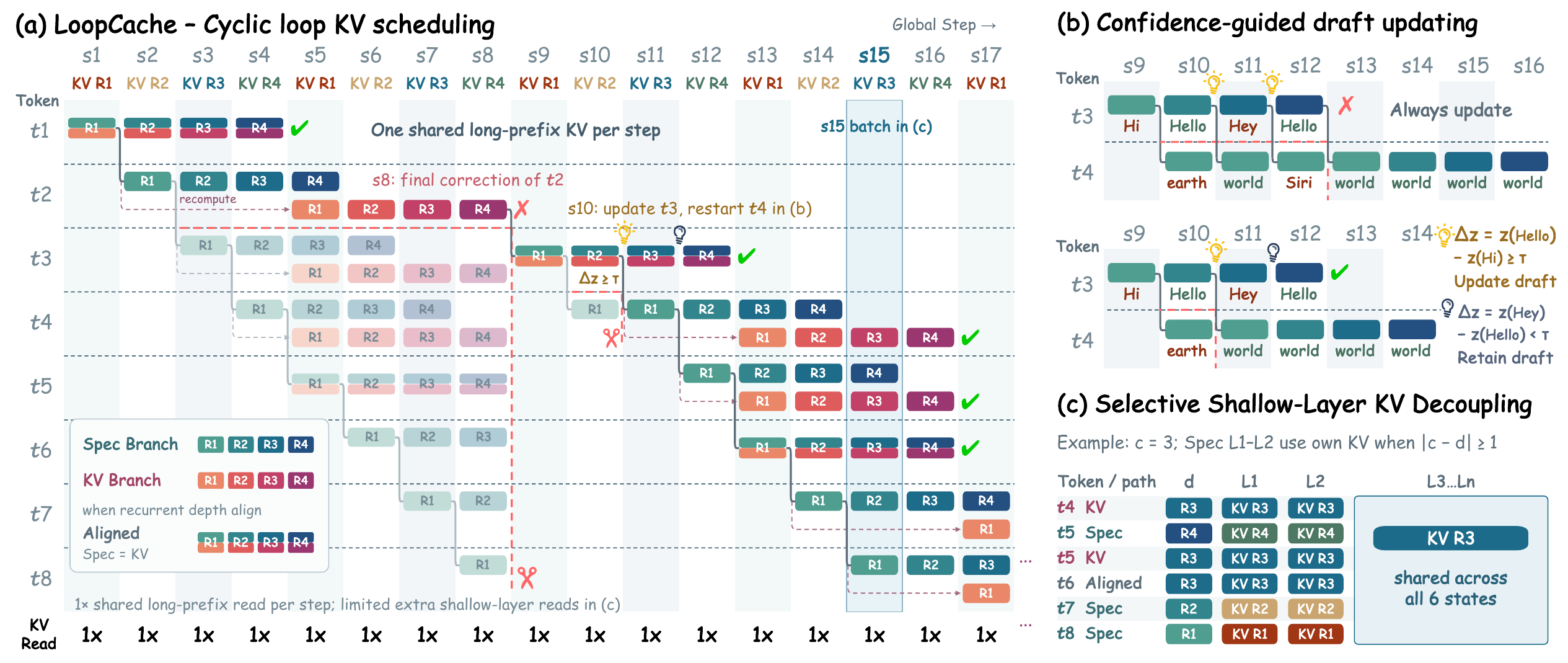}
    \caption{Overview of \method{}. (a) Cyclic loop KV scheduling with concurrent speculative and exact verification paths. (b) Confidence-guided dynamic draft updating versus unconditional updating. (c) Selective shallow-layer KV decoupling for the batch at $s_{15}$ in (a).}
    \label{fig:method}
\end{figure}

Based on this correlation, \method{} updates the draft only when $\Delta z_r\geq\tau$, where $\tau\geq0$ is a \mbox{configurable confidence threshold.}
Once triggered, the new candidate replaces the current draft, and all speculative descendants conditioned on the previous candidate are pruned.
Otherwise, the existing trajectory is retained.
Fig.~\ref{fig:method}(b) illustrates how this confidence-guided mechanism avoids repeated reversals compared with unconditional updating, preserving speculative progress without introducing \mbox{additional candidate branches.}

\subsection{Cyclic Loop KV Cache Scheduling}

Wave-style decoding overlaps drafting and verification but requires concurrent access to multiple loop-specific KV caches, causing substantial memory traffic in long contexts~\cite{Ha2026WaveFrontDP,Cho2026LoopSpecPS}.
To retain pipelined execution without this overhead, \method{} exploits cross-loop KV similarity for speculative computation while preserving exact loop-specific KV accesses for verification.

\textbf{Cyclic KV Scheduling.}
Instead of accessing different loop KV caches within the same global step, \method{} fixes the accessed cache according to a cyclic schedule.
For global decoding step $g\geq1$, the selected KV loop ID is
\begin{equation}
c(g)=1+((g-1)\bmod R),
\label{eq:cyclic_kv}
\end{equation}
where $R$ is the total recurrence depth.
Thus, the global steps repeatedly access $\text{KV R}_1\rightarrow\text{KV R}_2\rightarrow\cdots\rightarrow\text{KV R}_R$.
All active token states in each step share the same long-prefix KV cache $\text{KV R}_{c(g)}$, reducing the number of distinct full-prefix KV accesses from up to $R$ to one per step.
To support this schedule while maintaining exact decoding, \method{} organizes token computation into two concurrent paths, as shown in Fig.~\ref{fig:method}(a).

\textbf{Spec Branch.}
The speculative path advances in-flight tokens through their respective recurrent loops while using the KV cache selected by the current global step.
Let $d$ denote a token's current recurrent loop ID.
Even when $d\neq c(g)$, the token uses $\text{KV R}_{c(g)}$ to compute its intermediate state and predict subsequent tokens.
Although this computation is approximate, the high similarity of KV states across loops allows it to provide useful draft predictions.
The Spec Branch therefore retains the early drafting capability of wave-style decoding while sharing long-prefix KV accesses across tokens.

\textbf{KV Branch.}
To ensure exact verification, each surviving speculative token is recomputed through the correct loop-specific KV caches.
The KV Branch starts at the nearest upcoming complete KV cycle and advances through loops $1$ to $R$ in order, using the corresponding KV cache at each loop.
This recomputation restores the exact hidden states and KV entries under the candidate prefix, even when the Spec Branch predicts the correct token.
The KV Branch provides exact target logits for greedy decoding and rejection sampling.
If a candidate is rejected, its dependent speculative states are discarded, and decoding resumes from the corrected token.

Intermediate predictions from both branches can trigger confidence-guided draft updates, while the KV Branch determines the committed output.
When the two branches have matching recurrent depths, their computations are identical and can be merged into a single path, denoted as \textit{Aligned} in Fig.~\ref{fig:method}(a).
This avoids redundant computation while preserving exact target verification.

\subsection{Selective Shallow-Layer KV Decoupling}

Although cyclic KV sharing reduces memory traffic, using a mismatched loop cache can degrade draft predictions, especially when the draft loop differs substantially from the current KV loop.
As observed in Fig.~\ref{fig:observation-3}(b), shallow layers exhibit larger cross-loop KV differences than deeper layers.
To address this imbalance, \method{} selectively restores loop-specific KV accesses for a few shallow layers in the Spec Branch.

Let $L$ denote the number of physical layers in the recurrent block, and let $\mathrm{KV}_{\ell,r}$ denote the valid prefix KV cache of physical layer $\ell$ at loop $r$.
Given a selected shallow-layer set $\mathcal{S}\subseteq\{1,\ldots,L\}$ and a loop-ID difference threshold $\delta\geq1$, the KV cache used by the Spec Branch is
\begin{equation}
\mathrm{KV}_{\ell}^{\mathrm{Spec}}(d,c)=
\begin{cases}
\mathrm{KV}_{\ell,d},
& \ell\in\mathcal{S},\ |d-c|\geq\delta,\\
\mathrm{KV}_{\ell,c},
& \text{otherwise},
\end{cases}
\label{eq:shallow_kv}
\end{equation}
where $d$ is the draft's recurrent loop ID and $c=c(g)$ is the globally scheduled KV loop ID.

As illustrated in Fig.~\ref{fig:method}(c), selecting the first two physical layers with $\mathcal{S}=\{1,2\}$ and $\delta=1$ allows mismatched drafts to access their own loop-specific KV caches in these layers.
All remaining layers continue sharing the current cyclic KV cache.
This selective decoupling improves draft quality with only a small number of additional shallow-layer KV reads.
The KV Branch always uses the correct loop-specific caches, preserving exact verification.

%% file: docs/6-experiment.tex
\section{Experiments}
\label{sec:experiments}

\subsection{Experimental Setup}

\paragraph{Models and Hardware}
We evaluate \method{} on six Looped Transformers, including Ouro-1.4B, Ouro-2.6B, Huginn-3.5B, and three Raven variants based on Llama-3.2, OLMo-2, and TinyLlama~\cite{Zhu2025ScalingLR,Geiping2025ScalingUT,McLeish2025TeachingPL}. Table~\ref{tab:model_config} summarizes their configurations and per-loop KV cache footprints at 32K.
All experiments use eight NVIDIA A100 GPUs, with context parallelism for longer contexts.

\begin{table}[htbp]
    \centering
    \caption{Model configurations and per-loop KV cache sizes at 32K.}
    \label{tab:model_config}
    \small
    \setlength{\tabcolsep}{5pt}
    \renewcommand{\arraystretch}{1.15}
    \begin{tabular*}{0.88\linewidth}{@{\extracolsep{\fill}}lcccccc@{}}
        \toprule
        \shortstack{Model\\Metric}
        & \shortstack{Ouro\\1.4B}
        & \shortstack{Ouro\\2.6B}
        & \shortstack{Huginn\\3.5B}
        & \shortstack{Raven\\Llama-3.2}
        & \shortstack{Raven\\OLMo}
        & \shortstack{Raven\\TinyLlama} \\
        \midrule
        Params (B) & 1.40 & 2.60 & 3.50 & 1.38 & 1.35 & 0.84 \\
        Loops      & 4    & 4    & 32   & 32   & 32   & 32   \\
        KV (GiB)   & 6.00 & 12.00 & 2.58 & 0.375 & 1.50 & 0.25 \\
        \bottomrule
    \end{tabular*}
\end{table}

\paragraph{Baselines and Benchmarks}
We compare \method{} against AR, DtV, DFlash, WFD, and LoopSpec~\cite{Leviathan2022FastIF,Chen2026DFlashBD,Ha2026WaveFrontDP,Cho2026LoopSpecPS}. For DFlash, we train a block diffusion drafter using multiple datasets~\cite{Cobbe2021TrainingVT,codealpaca,Hendrycks2021MeasuringMP,Suzgun2022ChallengingBT,Bai2024LongAlignAR,Xu2024ChatQA2B}. Training supports sequence lengths up to 128K, with an effective batch size of 32 and a block size of 16. We evaluate context lengths from 4K to 128K on LongBench v1/v2~\cite{Bai2023LongBenchAB,Bai2024LongBenchVT}, RULER~\cite{Hsieh2024RULERWT}, and LongReason~\cite{Ling2025LongReasonAS}. For contexts beyond the native window, we apply YaRN for context extension~\cite{Peng2023YaRNEC}.

\subsection{Main Results}

\begin{figure}[!t]
    \centering
    \includegraphics[width=\linewidth]{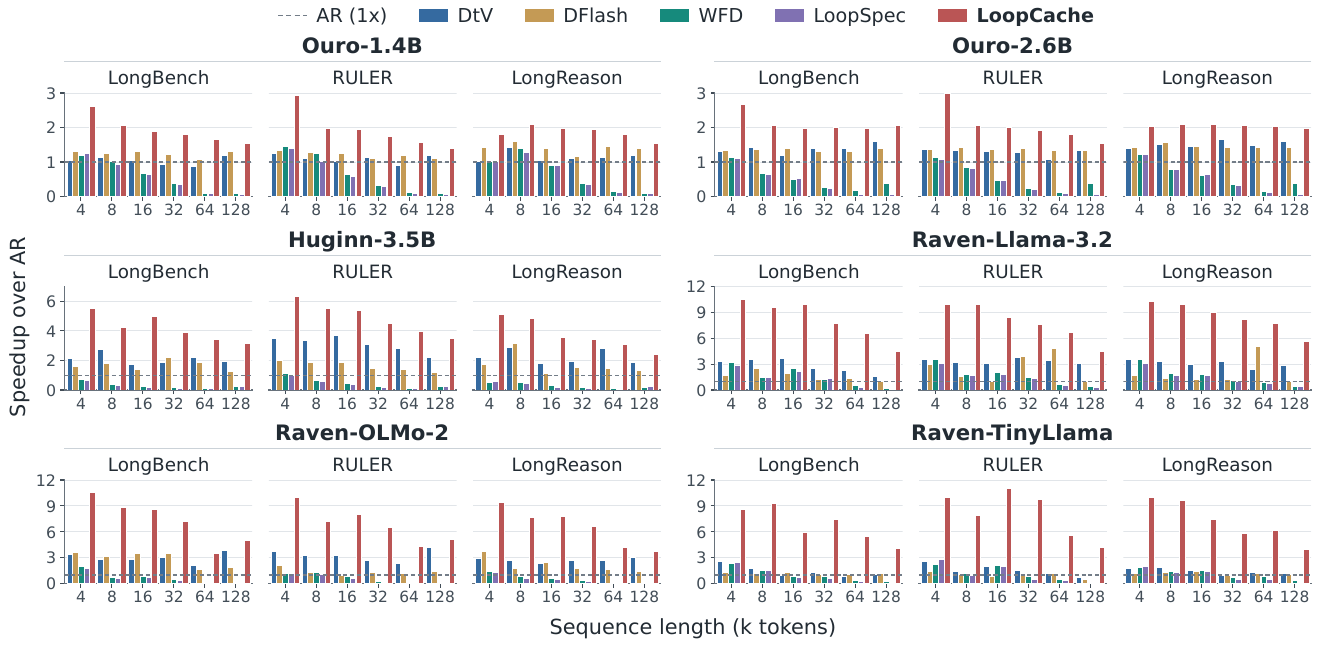}
    \caption{Decoding speedup across six Looped Transformers, three long-context benchmarks, and sequence lengths from 4K to 128K.}
    \label{fig:experiment-1}
\end{figure}

\paragraph{End-to-End Decoding Speedup}
Fig.~\ref{fig:experiment-1} presents the decoding speedup across different models, benchmarks, and context lengths. On average, \method{} achieves 10.25$\times$ and 14.56$\times$ speedup over WFD and LoopSpec, respectively, and 5.12$\times$ and 2.52$\times$ over AR and DtV. These results demonstrate the effectiveness of \method{} across diverse models and long-context workloads.

\begin{figure}[htbp]
    \centering
    \begin{minipage}[t]{0.485\linewidth}
        \vspace{0pt}
        {\centering\includegraphics[height=0.12\textheight]{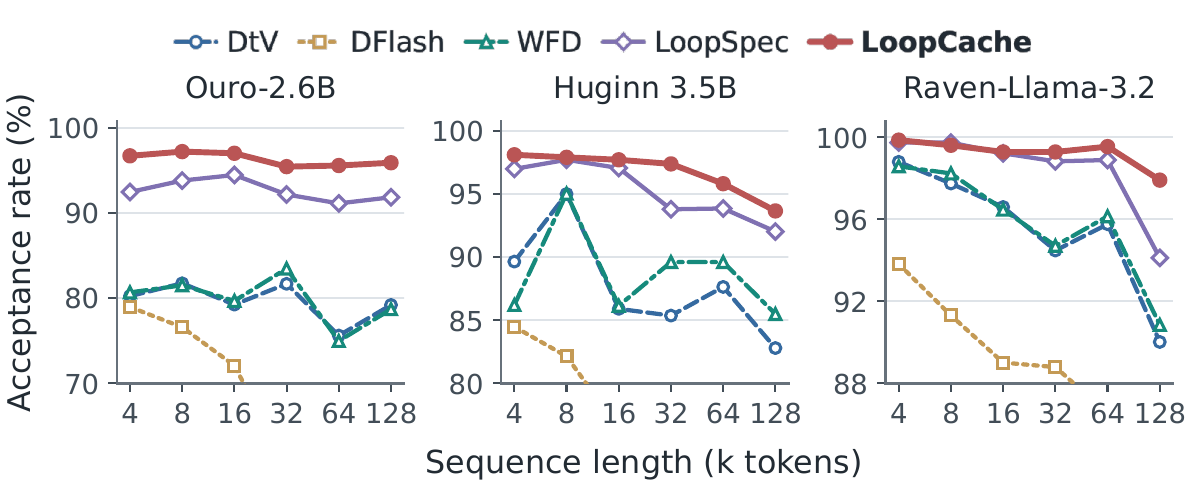}\par}
        \leftcaption{Acceptance rates of different \mbox{methods across models.}}
        \label{fig:experiment-2}
    \end{minipage}\hfill
    \begin{minipage}[t]{0.485\linewidth}
        \vspace{0pt}
        {\centering\includegraphics[height=0.12\textheight]{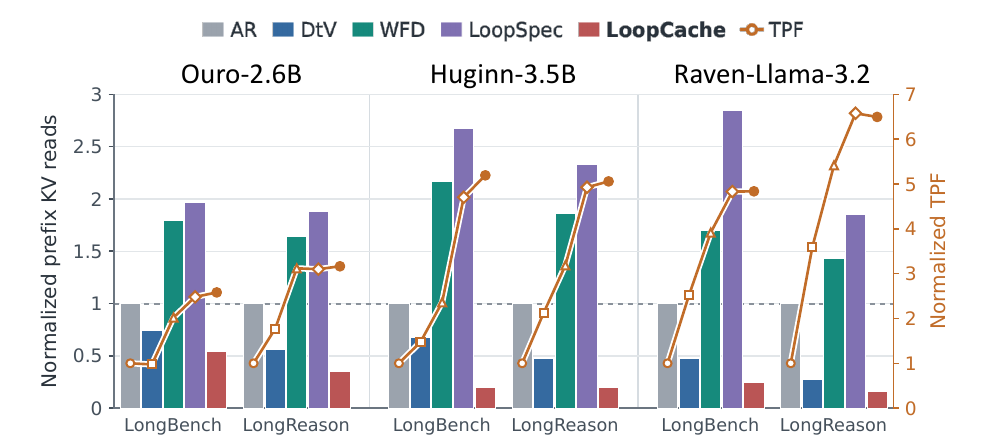}\par}
        \leftcaption{AR-normalized KV \mbox{cache reads and TPF.}}
        \label{fig:experiment-3}
    \end{minipage}
\end{figure}

\paragraph{Acceptance Rate}
Fig.~\ref{fig:experiment-2} compares the acceptance rates of different methods. \method{} maintains consistently high acceptance rates across different sequence lengths. Although \method{} uses approximate cross-loop KV states in most Spec Branch layers, confidence-guided dynamic draft updating corrects outdated predictions, while selective shallow-layer KV decoupling preserves critical loop-specific information. Together, these mechanisms enable \method{} to achieve higher acceptance rates than existing methods.

\paragraph{KV Cache Traffic and Token Parallelism}
We measure total KV cache reads and tokens per forward (TPF), where TPF is defined as the number of committed output tokens divided by the total number of loop forwards (global steps).
Fig.~\ref{fig:experiment-3} reports both metrics normalized to AR. While WFD and LoopSpec achieve high TPF, they incur substantial KV cache reads by accessing multiple loop-specific caches within each global step. 
In contrast, \method{} reads only one full-prefix loop-specific KV cache per global step, with limited additional shallow-layer accesses, substantially reducing memory traffic while maintaining high TPF.

\smallskip
\noindent
\begin{minipage}[t]{0.56\linewidth}
    \vspace{0pt}
    {\centering\includegraphics[width=0.94\linewidth]{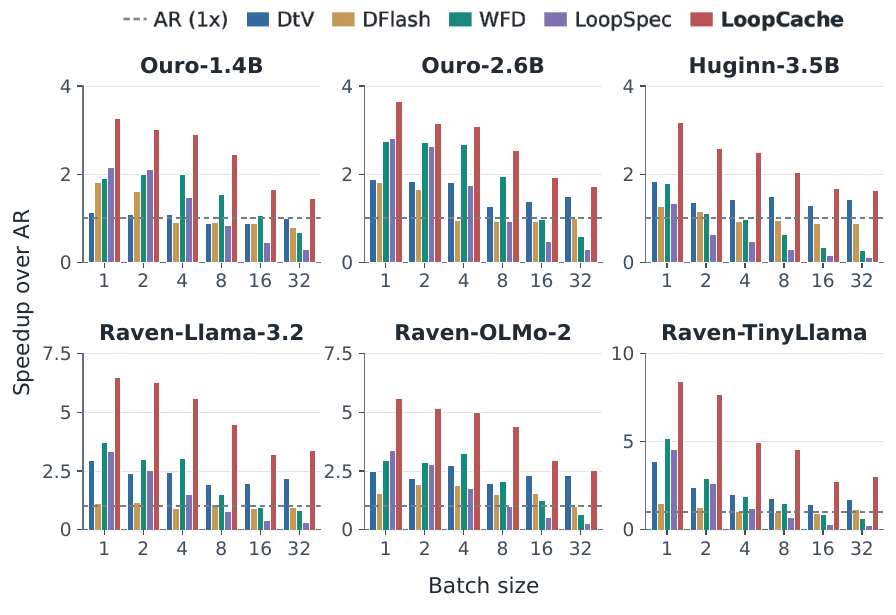}\par}
    \captionof{figure}{Decoding speedup at 4K across batch sizes from 1 to 32.}
    \label{fig:experiment-4}

    \paragraph{Batch-Size Scaling}
    Fig.~\ref{fig:experiment-4} shows the decoding speedup at a 4K context length with batch sizes ranging from 1 to 32 across models. Existing wave-style methods suffer from reduced speedups as the batch size increases, since KV cache traffic becomes increasingly dominant. By limiting KV cache accesses, \method{} maintains better performance across batch sizes.
\end{minipage}\hfill
\begin{minipage}[t]{0.42\linewidth}
    \vspace{0pt}
    \centering
    \includegraphics[width=0.97\linewidth]{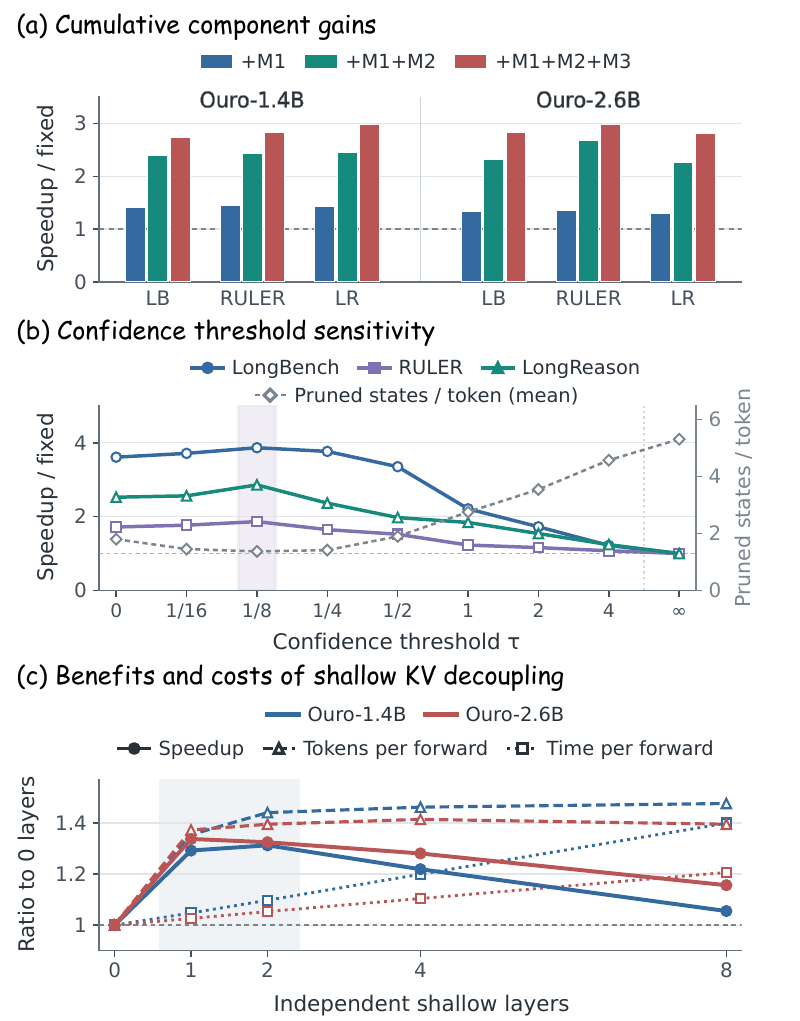}
    \captionof{figure}{Ablation study of \method{}. (a) Component contributions. (b) Confidence threshold. (c) Number of decoupled shallow layers.}
    \label{fig:experiment-5}
\end{minipage}\par

\subsection{Ablation Study}

Fig.~\ref{fig:experiment-5} evaluates the key design choices in \method{}.

\paragraph{Component Contributions}
Fig.~\ref{fig:experiment-5}(a) shows the cumulative speedup gains of the three components, normalized to the fixed-proposal baseline WFD.
M1 denotes confidence-guided draft updating, M2 denotes cyclic loop KV scheduling, and M3 denotes selective shallow-layer KV decoupling. 
Each component contributes additional gains, with the complete design achieving the best performance.

\paragraph{Confidence Threshold}
Fig.~\ref{fig:experiment-5}(b) evaluates the impact of the confidence threshold $\tau$. Pruned states per token measures the number of live speculative token states invalidated by draft replacement or verification rejection, normalized by the number of committed decoding tokens. We select $\tau=0.125$, which provides the best overall trade-off between speedup and speculative state pruning.

\paragraph{Shallow-Layer KV Decoupling}
Fig.~\ref{fig:experiment-5}(c) evaluates different numbers of decoupled shallow layers in terms of speedup, TPF, and time per forward, which denotes the latency of one loop forward. 
All metrics are normalized to the setting without shallow-layer decoupling. 
Selecting the first two physical layers as shallow layers preserves high TPF while limiting additional KV accesses and forward latency, achieving the best overall speedup.
Accordingly, we set $\mathcal{S}=\{1,2\}$ and the loop-ID difference threshold to $\delta=1$.

%% file: docs/7-conclusion.tex
\section{Conclusion}
\label{sec:conclusion}

We presented \method{}, an efficient self-speculative decoding framework for long-context Looped Transformers.
By exploiting intermediate-loop predictions, confidence-guided dynamic draft updating corrects outdated drafts early and avoids unnecessary speculative computation.
Cyclic loop KV scheduling substantially reduces memory traffic by sharing loop-specific KV accesses while preserving exact target verification.
Selective shallow-layer KV decoupling further improves draft acceptance with minimal additional overhead.
Extensive experiments demonstrate that \method{} achieves average speedups of 5.12$\times$ and 2.52$\times$ over AR and DtV, and 10.25$\times$ and 14.56$\times$ over WFD~\cite{Ha2026WaveFrontDP} and LoopSpec~\cite{Cho2026LoopSpecPS}, respectively.
These results highlight the effectiveness of \method{} in enabling fast and lossless long-context decoding for Looped Transformers.